\documentclass[conference]{IEEEtran}
\IEEEoverridecommandlockouts
\usepackage{cite}
\usepackage{amsmath,amssymb,amsfonts}
\usepackage{algorithmic}
\usepackage{graphicx}
\usepackage{textcomp}
\usepackage{xcolor}
\usepackage[numbers]{natbib}

\usepackage{bm}
\usepackage{algorithm}
\usepackage{algorithmic}
\usepackage{expl3}
\usepackage{xspace}
\ExplSyntaxOn
\newcommand{\latinabbrev}[1]{
  \peek_meaning:NTF .
  {#1\xspace}
  {
    #1.\xspace
  }
}
\ExplSyntaxOff

\graphicspath{{figures/}}

\def\eg{\latinabbrev{e.g}}

\def\ie{\latinabbrev{i.e}}

\newcommand{\E}{\mathrm{E}}
\newcommand{\SE}{\mathrm{SE}}

\newcommand{\ABS}{\mathrm{ABS}}
\newcommand{\std}{\mathrm{std}}
\newcommand{\MnU}{M\&U}

\def\BibTeX{{\rm B\kern-.05em{\sc i\kern-.025em b}\kern-.08em
    T\kern-.1667em\lower.7ex\hbox{E}\kern-.125emX}}
\begin{document}

\title{Magnitude and Uncertainty Pruning Criterion for Neural Networks}

\author{\IEEEauthorblockN{Vinnie Ko}
\IEEEauthorblockA{\textit{University of Oslo}\\
vinniebk@math.uio.no}
\and
\IEEEauthorblockN{Stefan Oehmcke}
\IEEEauthorblockA{\textit{University of Copenhagen}\\
stefan.oehmcke@di.ku.dk}
\and
\IEEEauthorblockN{Fabian Gieseke}
\IEEEauthorblockA{\textit{University of Copenhagen}\\
fabian.gieseke@di.ku.dk}
}
\IEEEpubid{978-1-7281-0858-2/19/\$31.00~\copyright~2019 IEEE}

\maketitle

\begin{abstract}
Neural networks have achieved dramatic improvements in recent years and depict the state-of-the-art methods for many real-world tasks nowadays. One drawback is, however, that many of these models are overparameterized, which makes them both computationally and memory intensive. Furthermore, overparameterization can also lead to undesired overfitting side-effects. Inspired by recently proposed magnitude-based pruning schemes and the Wald test from the field of statistics, we introduce a novel magnitude and uncertainty~(\MnU) pruning criterion that helps to lessen such shortcomings. One important advantage of our \MnU~pruning criterion is that it is scale-invariant, a phenomenon that the magnitude-based pruning criterion suffers from. In addition, we present a ``pseudo bootstrap'' scheme, which can efficiently estimate the uncertainty of the weights by using their update information during training. Our experimental evaluation, which is based on various neural network architectures and datasets, shows that our new criterion leads to more compressed models compared to models that are solely based on magnitude-based pruning criteria, with, at the same time, less loss in predictive power.
\end{abstract}


\section{Introduction}
In recent years, deep neural networks have achieved state-of-the-art performance for a broad range of tasks and have, hence, gained big popularity in many fields like computer vision and natural language processing~\citep{Goodfellow2016}.
Typically, these powerful models have millions or even billions of parameters, which require significant computational and memory resources when deployed. For the training phase, this computational bottleneck can usually be overcome by using powerful hardware such as graphics processing units, which renders it possible to train large models based on a lot of data in a reasonable amount of time. However, during the inference phase, once the trained model has been deployed, one is often faced with very limited computational resources. For instance, mobile or edge computing devices have little to no hardware acceleration available. In addition, many applications of neural networks---such as face recognition in video and autonomous driving---require the inference to be done with minimal latency. 

A prominent approach to deal with these scenarios is \emph{pruning}. Here, one usually generates a powerful model without any restrictions on its size and on its computational requirements during the training phase. Afterwards, one ``compresses'' the model such that it fits into the (main) memory of the device that should be used for the inference phase. Pruning neural networks this way can be a critical task for many real-world applications. In addition,~\cite{Denil2013} show that there is significant redundancy in the parameterization of many deep learning models. This overparameterization not only requires unnecessary computational resources, but can also lead to an overfitting of the models. Various approaches have been suggested in the literature that aim at decreasing the size and the computational footprint of neural networks~\citep{Han2015NIPS, Louizos2018, Chen2015, Srinivas2015, Ullrich2017}; see~\cite{Cheng2017} for an overview. One of the most successful works for pruning neural networks is provided by~\cite{Han2015NIPS}, who prune network parameters based on their magnitude, measured as absolute value.

\emph{Contribution:} Inspired by the \emph{Wald test} from the field of statistics~\citep{Silvey1959}, we introduce a novel magnitude and uncertainty~(\MnU) pruning criterion, which depicts a generalization of both, the magnitude based criterion used by~\cite{Han2015NIPS} and the Wald test statistic. One of the biggest advantages above the magnitude based pruning criteria~\citep{Han2015NIPS, Han2016ICLR, Guo2016, Zhu2017, Luo2017, Wen2016, Li2017} is that our \MnU~pruning criterion is scale-invariant. Furthermore, it can be easily applied to many papers on pruning strategies that exploit a magnitude based criterion. Furthermore, we also propose a new ``pseudo bootstrap'' scheme that efficiently estimates the weight uncertainties based on update information obtained during the training process. Our experiments show that using our \MnU~pruning criterion outperforms the previous pruning criterion and results in compressed models with less loss in predictive power.

\IEEEpubidadjcol
\section{Related Work}
The attempt to prune neural networks started with~\cite{LeCun1990} and~\cite{Hassibi1993}, who used a second-order Taylor expansion to determine the saliency of each parameter. Their methods require the computation of (the inverse of) the Hessian matrix regarding the loss function.~\cite{Anders1999} consider pruning as a statistical model selection problem and apply diverse model selection tools from the field of maximum likelihood theory.

Since the recent success of deep learning, there has been a number of new approaches for reducing the computation and memory footprint of neural networks. One approach, quantization, aims to reduce the numerical precision required to represent each parameter (or neuron). For example,~\cite{Vanhoucke2011},~\cite{Hwang2014}, and~\cite{Gong2014} compress networks by using 8-bit precision, 3-bit precision, and vector quantizations, respectively. There are more quantization based methods that expanded on these approaches~\citep{Courbariaux2015, Hubara2017, Lin2016, Rastegari2016}. Other strategies to reduce the computational burden resort to low-rank decompositions of tensors~\citep{Denton2014, Lebedev2014, Jaderberg2014}. In another category,~\cite{Hinton2015} uses distilling approach to obtain a small `student' network that can mimic what the bigger `teacher' network does.

The strategy that has received the most attention in recent years is probably the pruning of parameters based on certain criteria. Particularly,~\cite{Han2015NIPS} achieve state-of-the-art results by pruning weights based on their magnitudes (\ie, based on the absolute values of the parameters) and by retraining the network afterwards. The same authors further improve on their results by combining their absolute value based pruning scheme with quantizing and Huffman encoding techniques~\citep{Han2016ICLR}.
Extensions of this pruning strategy have also been proposed for neural machine translation~(NMT)~\citep{See2016} and for recurrent neural networks~(RNN)~\citep{Narang2017}, respectively. 
Yet, those two follow-up works keep the same criterion as the one originally proposed by~\cite{Han2015NIPS} for ranking the weights to be pruned, \ie, they use the absolute values of the weights. Finally, there are also attempts to prune specific classes of weights~\citep{Anwar2017, Changpinyo2017, Lebedev2016, Li2017}. These approaches have the advantage that the resulting pruned networks have less irregular network connections and are, hence, better suited for parallel implementations.

\section{Background}\label{section:background}
In this section, we briefly recap the theory behind statistical hypothesis testing and look into the similarities between the Wald test statistic and the magnitude based criterion~\citep{Han2015NIPS}.

\subsection{Asymptotic Normality}\label{section:asymptotic_normality}
Consider a regression model $g$ with independent and dependent variable pairs ${(\bm{x}_{1},\bm{y}_{1}), \dots, (\bm{x}_{n},\bm{y}_{n})}$. Further, let $\bm{X} = [\bm{x}_{1}, \dots, \bm{x}_{n}]^{\rm T}$ denote a tensor containing all~$n$ observations of independent variables. The arbitrary regression model $g$ can then be written as 
\begin{align}
\E[\bm{Y}] = g(\bm{X}, \bm{w}),
\end{align}
where $\bm{w} = [w_{1}, \dots, w_{p}]^{\rm T}$ are the parameters of the model, which are commonly called \emph{weights} in case of neural networks. The parameter $p$ indicates the total number of parameters of the model.

For now, we assume that there exists a unique set of optimal parameters $\bm{w}_{\star}$, which will give the best approximation of $\bm{y}$. We can then obtain $\widehat{\bm{w}}$, our estimate of $\bm{w}_{\star}$, by minimizing a loss function with the optimization technique of choice (\eg, stochastic gradient descent, mini-batch gradient descent, momentum based methods, etc.).

When the loss function is the negative log-likelihood, or an equivalent quantity such as cross entropy, gradient descent can be seen as a numerical realization of maximum likelihood~(ML) estimation. In this case, we can apply the results from classic literature on ML theory~\citep{Casella2002, LeCam1990, White1982} and we have for $n \to \infty$, the following convergence in distribution property:
\begin{align}\label{eq:ML_asymptotic_normality}
\small
\sqrt{n}\left(\widehat{\bm{w}} - \bm{w}_{\star}\right) 
~\overset{d}{\to}~\mathcal{N}(\bm{0}, \bm{V}_{\mathrm{ML}}).
\end{align}
Here $\bm{V}_{\mathrm{ML}}$ is the covariance matrix as defined in ML theory. In the field of statistics, this type of convergence to a normal distribution is called \emph{asymptotic normality}~\citep{vanderVaart2000}. In the model robust case, $\bm{V}_{\mathrm{ML}}$ is equal to the so-called sandwich estimator
\begin{align}\label{eq:V_ML_model_robust}
\small
\bm{V}_{\mathrm{ML}} = \bm{\mathcal{I}}_{\bm{w}}^{-1} \bm{K}_{\bm{w}} \, \bm{\mathcal{I}}_{\bm{w}}^{-1},
\end{align}
where $\bm{\mathcal{I}}_{\bm{w}}$ is the Fisher information and where $\bm{K}_{\bm{w}}$ is the covariance matrix of the score function~\citep{White1982}. In statistics, the score function is defined as the first derivative of the log-likelihood function with respect to its parameters. It indicates how sensitive a likelihood function is w.r.t.\ its parameters. For exact definitions of the quantities in Equation~\eqref{eq:V_ML_model_robust}, we refer to the literature in maximum likelihood theory, for instance by~\cite{White1982}. When one assumes that the model is true, in other words, that the current candidate model is the true model that generated the data, we have $\bm{K}_{\bm{w}} = \bm{\mathcal{I}}_{\bm{w}}$ and Equation~\eqref{eq:V_ML_model_robust} simplifies to
\begin{align}\label{eq:V_ML}
\small
\bm{V}_{\mathrm{ML}} = \bm{\mathcal{I}}_{\bm{w}}^{-1}.
\end{align}
For more details on the impact of this `true model' assumption and the comparison between Equations~\eqref{eq:V_ML_model_robust} and \eqref{eq:V_ML}, we refer to the corresponding literature~\citep{Hardin2003, Kauermann2000, Aauermann2001}.

Even though stochastic gradient descent~(SGD) and its modified versions can be considered as numerical realization of ML estimation, the details of SGD lead to properties different from the ones obtained via ML. The recent work by~\cite{Toulis2017} obtains, for the first time, a full analytical characterization of the asymptotic behavior of SGD procedures, including an asymptotic normality and analytical expression for the covariance matrix, which is different from that of ML estimators. They also show the loss of asymptotic statistical efficiency for SGD estimators. Although the work of~\cite{Toulis2017} is a great step forward towards understanding the asymptotic properties of SGD estimators, its result cannot be directly applied to non-vanilla versions of SGD without necessary extension work. For example, combining the SGD estimator with training strategies like Dropout~\citep{Srivastava2014} and Dropconnect~\citep{Wan2013} or using momentum based modifications such as RMSprop~\citep{Tieleman2012} or Adam~\citep{Kingma2015} would invalidate the asymptotic covariance matrix from~\cite{Toulis2017}.

Based on~\cite{White1982} and~\cite{Toulis2017}, it is not unreasonable to assume that the modified versions of the SGD estimator are still consistent and have asymptotic normality with a finite, but unknown covariance matrix $\bm{V}$, \ie, that we have
\begin{align}\label{eq:SGD_asymptotic_normality}
\sqrt{n}\left(\widehat{\bm{w}} - \bm{w}_{\star}\right) 
~\overset{d}{\to}~\mathcal{N}(\bm{0}, \bm{V}).
\end{align}

\subsection{Statistical Hypothesis Testing}\label{section:statistical_hypothesis_testing}
With the asymptotic normality property as defined in Section~\ref{section:asymptotic_normality}, one can perform statistical hypothesis testing on the parameters $\bm{w}$. The natural null hypothesis would be
\begin{align}\label{eq:null_hypothesis}
H_{0}: w_{j} = 0,
\end{align}
\ie, is the $j$-th parameter of the model significantly different from 0? The three commonly used statistical hypothesis testings are the \emph{Wald test}~\citep{Silvey1959}, the \emph{likelihood ratio test}~\citep{Neyman1933}, and the \emph{score test}~\citep{Bera2001}. Note that all three tests are asymptotically equivalent~\citep{Engle1984}. Yet, the Wald test has a practical advantage over the other two since it does not require nested models for testing the null hypothesis on individual weights. Here, ``nested'' means that one can obtain one model by constraining some of the parameters of another model.

Given Equations~\eqref{eq:SGD_asymptotic_normality} and~\eqref{eq:null_hypothesis}, the Wald test statistic for the parameter $w_{j}$ is given by
\begin{align}\label{eq:wald_test_statistic}
\small
z = \frac{\widehat{w}_{j}}{\SE(\widehat{w}_{j})} = \frac{\widehat{w}_{j}}{\sqrt{\bm{\widehat{V}}_{j,j}}},
\end{align}
where $\SE(\widehat{w}_{j})$ indicates the standard error of $\widehat{w}_{j}$ and $\bm{\widehat{V}}$ the estimate of the covariance matrix from Equation~\eqref{eq:SGD_asymptotic_normality}.\footnote{Since we defined $\bm{V}$ as a finite, but unknown covariance matrix, its estimate $\bm{\widehat{V}}$ is an unknown and arbitrary estimate.} The test statistic $z$ follows the standard normal distribution~\citep{Engle1984}. The corresponding $p$-value can be obtained via $p = 2(1-\Phi(|z|))$, where $\Phi$ is the cumulative distribution function of the standard normal distribution.

\subsection{Wald Test and Absolute Value Based Pruning}\label{section:relationship_Wald_ABS}
In statistics, one of the most canonical ways of removing redundant parameters is by performing null hypothesis testing as described above and by dropping the parameters that have a $p$-value above a certain threshold (\eg, 0.05 or 0.01). Since $p = 2(1-\Phi(|z|))$ is a monotonic function of $|z|$, one can skip the step of converting $|z|$ into a $p$-value. Instead, one can directly use
\begin{align}\label{eq:wald_test_statistic_absolute}
\small
|z| = \frac{|\widehat{w}_{j}|}{\SE(\widehat{w}_{j})} = \frac{|\widehat{w}_{j}|}{\sqrt{\bm{\widehat{V}}_{j,j}}},
\end{align}
and can compare it with the threshold values in $|z|$-scale, which can be obtained by transforming the threshold $p$-values via the function $\Phi^{-1}(1-p/2)$. Here, we can realize two things. Firstly, Equation~\eqref{eq:wald_test_statistic_absolute} is highly similar to the magnitude based pruning criterion suggested by~\cite{Han2015NIPS} and also used by~\cite{Han2016ICLR},~\cite{See2016}, and~\cite{Narang2017}. In those works, only the absolute value of each parameter is considered and the parameters with absolute values below a certain threshold are removed. We denote this criterion as the $\ABS$ pruning criterion from now on. The only difference between the Wald test statistic (Equation~\eqref{eq:wald_test_statistic_absolute}) and the $\ABS$ pruning criterion is the standard error term in the denominator.

In this sense, the Wald test statistic can actually be interpreted as evaluating parameters based on their absolute value, while compensating for their uncertainty. For example, if $\widehat{w}_{j}$ has high uncertainty, there is higher chance that we observe a large value of $\widehat{w}_{j}$ ``by chance''. Dividing by the standard error compensates for this uncertainty.

\section{Magnitude and Uncertainty Pruning Criterion}\label{section:pruning_criterion}
Inspired by Wald test, we evaluate the importance of parameters by taking both their magnitudes and their uncertainties into account. By using this combined criterion, we can prune to the desired proportion of parameters in a model, specific layer, feature map, or any selected part of a model. In addition, we also introduce a ``pseudo bootstrap'' approach to efficiently estimate the uncertainties of the weights by keeping track of their changes during the training process.

\subsection{Definition}\label{section:definition}
Based on our observation of the similarity between the Wald test statistic and the $\ABS$ pruning criterion in Section~\ref{section:relationship_Wald_ABS} as well as based on the interpretation of the Wald test statistic, we define our \MnU~pruning criterion as
\begin{align}\label{eq:MU_pruning_criterion}
\tau_{j} = \frac{|\widehat{w}_{j}|}{\lambda + \widetilde{\sigma}_{\widehat{w}_{j}}}.
\end{align}
Here, $\widetilde{\sigma}_{\widehat{w}_{j}}$ is an uncertainty estimate of $\widehat{w}_{j}$, a quantity that should reflect how uncertain the estimated parameter $\widehat{w}_{j}$ is. We will cover possible choices for $\widetilde{\sigma}_{\widehat{w}_{j}}$ in Sections~\ref{section:how_to_obtain_SE} and~\ref{section:pseudo_bootstrap}. The hyperparameter $0 \leq \lambda$ is a user-defined parameter that determines the balance between magnitude and uncertainty. When $\lambda \to \infty$, no uncertainty is taken into account and the \MnU~pruning criterion becomes equal to the $\ABS$ pruning criterion used by~\cite{Han2015NIPS}. As $\lambda$ decreases, more and more uncertainty is taken into account. In case of $\lambda = 0$, the criterion $\tau_{j}$ has the same functional form as the Wald test statistic. If one chooses $\widetilde{\sigma}_{\widehat{w}_{j}}$ such that it is an analytically justified estimate of the standard error, $\tau_{j}$ is simply the Wald test statistic. This implies that \MnU~pruning criterion can be seen as a generalization of the Wald test statistic and the $\ABS$ pruning criterion.

\subsection{Scale Invariant Property}\label{section:scale_invariant_property}
Pruning strategies that are based on the magnitude based pruning criterion~\citep{Han2015NIPS, Han2016ICLR, See2016, Narang2017, Guo2016, Zhu2017, Luo2017, Wen2016, Li2017} suffer from the so-called scale issue by their nature.

Consider a very simple model (e.g. linear regression) with 2 input variables $X_{1}$ and $X_{2}$.  Assume that the weights are $w_{1} = 10$ and $w_{2} = 0.1$. One can be easily tricked to think that $X_{1}$ is more important than $X_{2}$. However, if we divide $X_{2}$ by 1000 (e.g. m to km), $w_{2}$ will get much bigger and this will change which weight is pruned by the magnitude based pruning criterion. In contrast, akin to the Wald test statistic, our \MnU~pruning criterion cancels out this ‘change of scale effect’ by standardizing with $\widetilde{\sigma}_{\widehat{w}_{j}}$. Moreover, we introduce in Section~\ref{section:reparameterization_of_lambda} a reparametrization trick that makes the hyperparameter $\lambda$ robust to the change of scale.

\subsection{Choosing Uncertainty Estimates}
\label{section:how_to_obtain_SE}
The first thing to note is that $\widetilde{\sigma}_{\widehat{w}_{j}}$ is not the standard error, but concerns a much more general concept of uncertainty estimate. Ideally, the standard error is a very suitable candidate as an uncertainty estimate $\widetilde{\sigma}_{\widehat{w}_{j}}$, like in the original Wald test statistic formula. However, there is at the moment no asymptotic theories developed for non-convex neural networks with commonly used estimation methods such as Adam~\citep{Kingma2015} or RMSprop~\citep{Tieleman2012}.

In rare cases, when the optimization task induced by a neural network is convex (or piecewise convex) with the negative log-likelihood, or an equivalent quantity such as cross entropy as loss function, one can take analytically derived asymptotic variance formulas such as Equation~\eqref{eq:V_ML_model_robust}, Equation~\eqref{eq:V_ML}, or other ones~\citep{Toulis2017} and estimate these uncertainty measures via their sample equivalent (\eg, by replacing the expectation with a sum over the samples). However, estimating those quantities has a number of numerical bottlenecks. For example, to obtain the sample equivalent of $\bm{\mathcal{I}}_{\bm{w}}$, one needs to calculate the Hessian matrix, which requires the second derivative of the log-likelihood function with respect to all the parameters in the model. Although modern deep learning software offers highly optimized automatic differentiation for the first derivative, many of the automatic differentiation implementations do not support the second derivative. In case they do support it, the corresponding implementations are not as highly optimized as those for the first derivative.

Another computational bottleneck is the computation of~$\bm{K}_{\bm{w}}$. Computation of this matrix requires the first derivative of the log-likelihood with respect to the model parameters, for every sample separately. This is highly inefficient for most deep learning software packages since optimized for mini-batch operations. Yet, the biggest computational obstacle is that one needs to take the inverse of $\bm{\mathcal{I}}_{\bm{w}}$. For a network with, say, a million parameters, inverting this matrix is infeasible in practice even using significant computational resources.

\subsection{Pseudo Bootstrap}\label{section:pseudo_bootstrap}
We introduce two possible alternatives to compute the uncertainty estimate~$\widetilde{\sigma}_{\widehat{w}_{j}}$. The first alternative is to make use of bootstrapping~\citep{Efron1994}. There have been extensive studies done on both theoretical and numerical properties of this approach~\citep{Bickel1981, Efron1992, Davison1997, Shao2012} and it is a canonical method in statistics for estimating the standard error in case an analytical estimate is not feasible. The disadvantage is that we need to train the same model multiple times, which is a big increase in the computational costs.\footnote{Since neural networks are usually trained with large number of observations, the Jackknife resampling method~\citep{Efron1982}, which requires repeating the training process equal to the number of total data points, is not suitable.}

The second alternative is a novel estimation scheme, called \emph{pseudo bootstrap}, which uses uncertainty information gathered during the training process. This idea stems from the fact that the training process with mini-batch SGD has similarities with the bootstrap resampling strategy: Each mini-batch we draw during training can be seen as a bootstrap sample with replacement and of smaller size. In addition, the iterative weight update process of mini-batch SGD can be seen as training with bootstrap samples, where each bootstrap training instance uses the parameter values from the previous instance as parameter initialization.

\begin{algorithm}[t]
\caption{\footnotesize \emph{Pseudo bootstrap}, our proposed novel algorithm for a fast estimation of the uncertainty of each weight in neural network. The algorithm described below computes a single weight $w_{j}$ and has to be repeated for all the weights in the neural network.}
\label{alg:pseudobootstrap}
\begin{algorithmic}[1]
\footnotesize
\REQUIRE{$B \in \mathbb{N}$: Weight collection size}
\REQUIRE{$\bm{w} = [w_{1}, \dots, w_{p}]$: Parameter vector of the given neural network.}
\REQUIRE{$n_{\mathrm{iter}}$: The total number of weight update iterations that is going to be made during the training phase. Note that this is the total number of iterations and not the total number of epochs.}
\STATE{$\bm{w}_{j,\{B\}}$: An empty vector of length $B$.}
\STATE{$i_{\mathrm{iter}} \leftarrow 0$ (Initialize training iteration count)}
\STATE{$i_{B} \leftarrow 0$ (Initialize weight collecting count)}

\WHILE{training the target model with mini-batch SGD}\label{while_loop_start}
\STATE{$i_{\mathrm{iter}} \leftarrow i_{\mathrm{iter}}+1$}
\IF{$n_{\mathrm{iter}} - B < i_{\mathrm{iter}} \leq n_{\mathrm{iter}}$}
\STATE{$i_{B} \leftarrow i_{B}+1$}
\STATE $\bm{w}_{j,\{B\}}[i_{B}] \leftarrow w_{j}$ (Record current value of $w_{j}$.)
\ENDIF
\ENDWHILE\label{while_loop_end}
\STATE{$\widetilde{\sigma}_{\widehat{w}_{j}} = \std(\bm{w}_{j,\{B\}})$} (Compute uncertainty estimate of $\widehat{w}_{j}$ by using its `fluctuations' during the training.)\label{std}
\STATE{\textbf{return} $\widetilde{\sigma}_{\widehat{w}_{j}}$}
\end{algorithmic}
\end{algorithm}

The pseudo bootstrap approach is sketched in Algorithm~\ref{alg:pseudobootstrap}: To obtain the uncertainty estimate $\widetilde{\sigma}_{\widehat{w}_{j}}$, the algorithm monitors how $\widehat{w}_{j}$ has changed during the last $B$ iterations of mini-batch SGD training and by storing that information in the `weight collection' $\bm{w}_{j,\{B\}} = [w_{j,1}, \dots, w_{j,B}]$ (see Lines~\ref{while_loop_start}--\ref{while_loop_end}). After the training, we simply obtain the uncertainty estimate via $\widetilde{\sigma}_{\widehat{w}_{j}} = \std(\bm{w}_{\{B\}})$ (see Line~\ref{std}). Note that other ``collecting strategies'' such as recording weight values through the entire training process with a certain interval between them have been tested. However, the best results were obtained by recording the weight values during the last iterations.

\begin{figure*}[h]
\begin{center}
\includegraphics[width=0.95\linewidth]{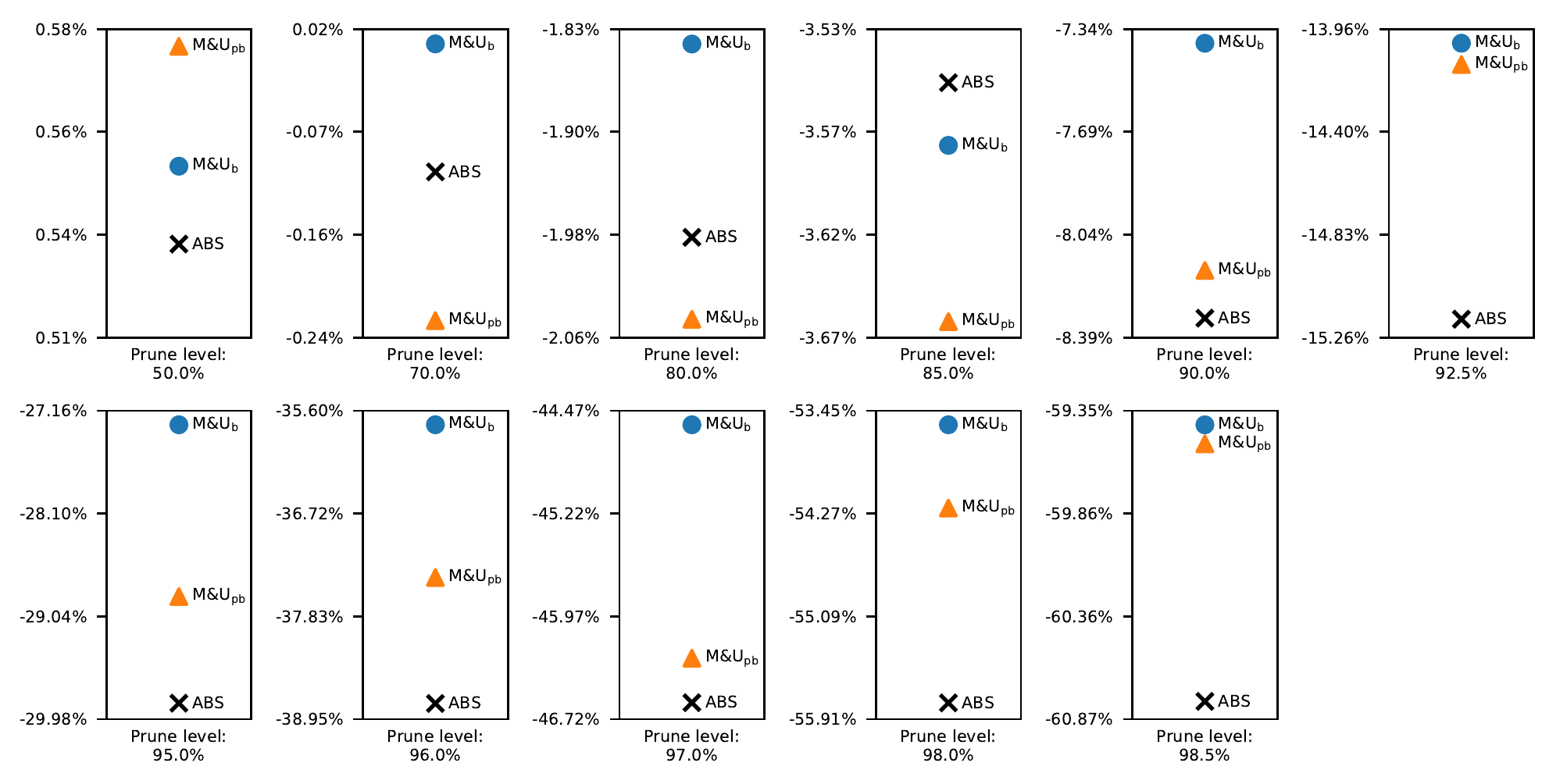}
\end{center}
\caption{\small Result of Experiment 1. The $y$-axis of each plot shows the loss in test accuracy due to pruning (with retraining afterwards). It was computed as the test accuracy of the pruned model subtracted by the test accuracy of the unpruned original model. Blue circles with `\MnU\textsubscript{b}' indicate that our \MnU~pruning method with $\widetilde{\sigma}_{\widehat{w}_{j}}^{\mathrm{boot}}$ was used. Orange triangles with `\MnU\textsubscript{pb}' mean that our \MnU~pruning method with $\widetilde{\sigma}_{\widehat{w}_{j}}^{\mathrm{pboot}}$ was used. Black crosses with `ABS' mean that the $\ABS$ pruning criterion from~\cite{Han2015NIPS} was used. The test accuracy of the unpruned model was $81.21\%$.}
\label{fig:FMNIST}
\end{figure*}

\subsection{Reparameterization of $\lambda$}\label{section:reparameterization_of_lambda}
The hyperparameter $\lambda$ determines the tradeoff between magnitude and uncertainty. It is therefore crucial to consider reasonable values for $\lambda$. Note that the actual impact of $\lambda$ on the balance between magnitude and uncertainty depends largely on which scale $w_{j}$ is. For instance, if we decide to use batch normalization and the absolute value of $w_{j}$ suddenly becomes much smaller, then a previously ``optimal'' value for $\lambda$ (\eg, $\lambda=1$) might not be optimal anymore. To deal with this problem, we suggest to reparametrize~$\lambda$ via 
\begin{align}\label{eq:lambda_reparameterization}
 \lambda = \lambda^{*} \cdot \std(\bm{w}_{\{\mathrm{layer}_{k}\}})
\end{align}
when one prunes within layer $k$. Here, $\lambda^*$ is a fixed global assignment for $\lambda$ and $\std(\bm{w}_{\{\mathrm{layer}_{k}\}})$ depicts the standard deviation of the weights within the $k$-th layer.\footnote{Naturally, instead of considering $\bm{w}_{\{\mathrm{layer}_{k}\}}$, one can also consider the standard deviation based on all the weights via $\bm{w}_{\{\mathrm{model}\}}$ such that $\lambda$ is scaled by the whole model (or by a specific part of a certain layer).}

\section{Experiments}
We implemented our \MnU~pruning criterion in PyTorch~\citep{Paszke2017} and evaluated our approach via four experiments that were based on the Fashion MNIST, CIFAR-10, MNIST, and ImageNet datasets, respectively. Experiment 1 was repeated 100 times, Experiment 2 and 3 were repeated 20 times, and Experiment 4 was repeated 10 times. The reported results are averaged results over those repetitions.

Since the majority of pruning strategies utilize the magnitude based pruning criterion, it is most informative to compare our method directly to the $\ABS$ criterion from~\cite{Han2015NIPS}. Our \MnU~pruning criterion can be easily extended to replace the pruning criterion part of, for example, automatic gradual pruning~\citep{Zhu2017}, structured pruning~\citep{Luo2017, Li2017}, dynamic network surgery~\citep{Guo2016} and more. Note further that our \MnU~pruning criterion should be seen as a companion rather than a competitor to other model compression approaches such as distilling \citep{Hinton2015} and quantization \citep{Vanhoucke2011}, since it can be combined with those techniques.

\begin{figure*}[h]
\begin{center}
\includegraphics[width=0.95\linewidth]{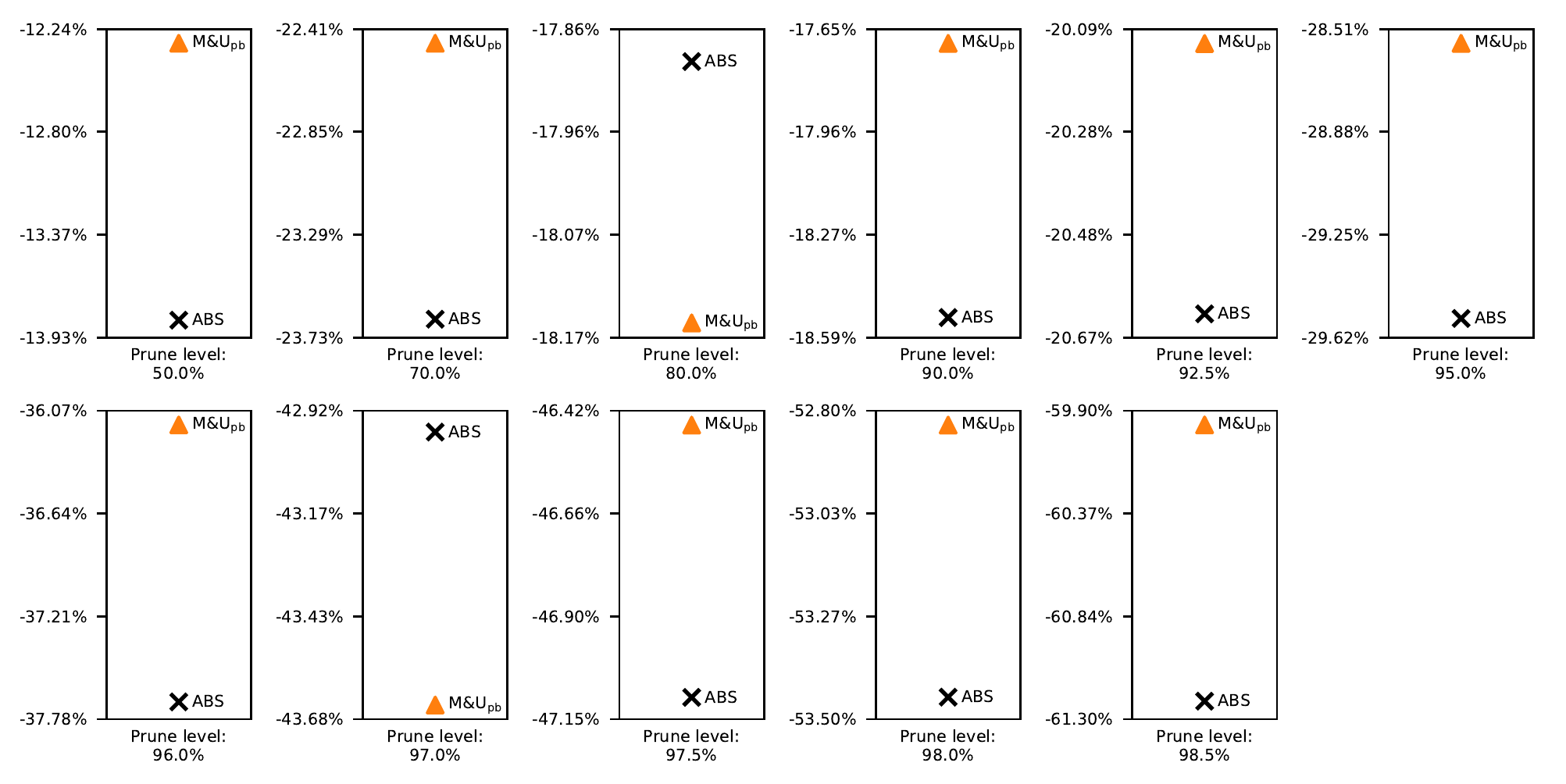}
\end{center}
\caption{\small Result of Experiment 2. The $y$-axis is the same as in Figure~\ref{fig:FMNIST}. Orange triangles with `\MnU\textsubscript{pb}' mean that our \MnU~pruning method with $\widetilde{\sigma}_{\widehat{w}_{j}}^{\mathrm{pboot}}$ was used. Black crosses with `ABS' mean that the $\ABS$ pruning criterion from~\cite{Han2015NIPS} was used. The test accuracy of the unpruned model was $79.98\%$.}
\label{fig:CIFAR10}
\end{figure*}

\begin{figure*}[h]
\begin{center}
\includegraphics[width=0.95\linewidth]{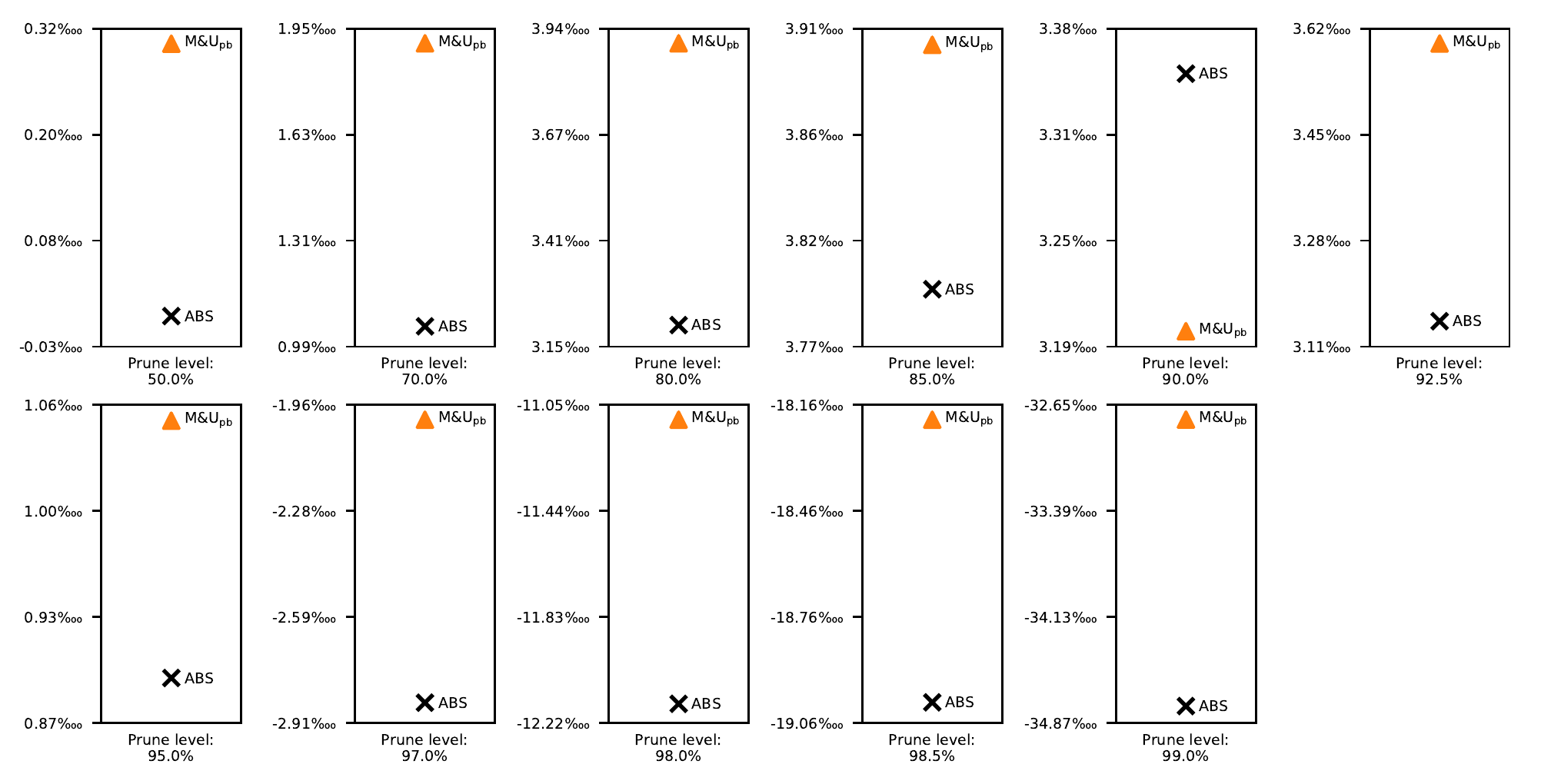}
\end{center}
\caption{\small Result of Experiment 3. The $y$-axis is the same as in Figure~\ref{fig:FMNIST}. Note that the unit of the $y$-axis is basis point, and not percent. Orange triangles with `\MnU\textsubscript{pb}' mean that our \MnU~pruning method with $\widetilde{\sigma}_{\widehat{w}_{j}}^{\mathrm{pboot}}$ was used. Black crosses with `ABS' mean that the $\ABS$ pruning criterion from~\cite{Han2015NIPS} was used. The test accuracy of the unpruned model was $98.5160\%$.}
\label{fig:MNIST}
\end{figure*}

\subsection{Experiment 1: Comparison of Uncertainty Estimates}\label{section:experiment_1}
The Fashion MNIST dataset consists of 28x28 grayscale images, each with a label from one of 10 classes~\citep{Xiao2017}. To make the bootstrap estimation computationally feasible, we kept the size of this experiment as small as possible. Thus, we only used a balanced subset of 6,000 examples from the training data. From this subset, 600 examples were set apart for validation during training. For testing, the entire test set was used. We employed a very small feedforward convolutional neural network (CNN) with two convolution layers and three fully connected layers, combined with max pooling, ReLU activation function and dropout. The first fully connected layer got a feature map of size 6 x 6 x 6 as input. The two hidden layers had 32 and 48 neurons, respectively, and the model had a total amount of 9,660 parameters. We chose cross entropy as loss function and the model was trained for 500 epochs via RMSprop.

Firstly, since the total number of parameters was small, we tried to estimate the covariance matrix of $\bm{w}$ with analytical expressions~\eqref{eq:V_ML_model_robust} and \eqref{eq:V_ML}. However, $\bm{\mathcal{I}}_{\bm{w}}$ and $\bm{K}_{\bm{w}}$ were not invertible most of the time. As alternative, we obtained the psuedo inverse of those matrices. Yet, the pseudo inverse contained almost always negative diagonal elements. As a sanity check, we estimated the expectation of the score vector and many of its elements had values that were far away from 0. This suggests that the maximum likelihood theory did not hold in this case. This was expected, since many of the assumptions from~\cite{LeCam1990} and~\cite{Toulis2017}, including convexity, were broken. Secondly, we estimated $\widetilde{\sigma}_{\widehat{w}_{j}}$ with bootstrap. We drew 100 bootstrap datasets and trained 100 independent instances of the model. With the weights from those 100 trained models, we obtained the bootstrap based uncertainty estimate $\widetilde{\sigma}_{\widehat{w}_{j}}^{\mathrm{boot}}$. Lastly, we obtained the pseudo bootstrap based uncertainty estimate $\widetilde{\sigma}_{\widehat{w}_{j}}^{\mathrm{pboot}}$, with the method described in Section~\ref{section:pseudo_bootstrap} and $B = 200$. We tried different values of $\lambda^{*}$ and empirically choose $\lambda^{*} = 10^{-1}$ for $\widetilde{\sigma}_{\widehat{w}_{j}}^{\mathrm{boot}}$ and $\lambda^{*} = 10^{-4}$ for $\widetilde{\sigma}_{\widehat{w}_{j}}^{\mathrm{pboot}}$.

With the obtained $\widetilde{\sigma}_{\widehat{w}_{j}}$ and $\lambda$, we pruned $a\%$ of parameters by using our \MnU~pruning criterion from each fully connected layer. The values of $a$ that were tested can be found in Figure~\ref{fig:FMNIST}. For comparison, we also pruned the model by using the $\ABS$ pruning criterion~\citep{Han2015NIPS}. After pruning, each model was retrained for 200 epochs.

Figure~\ref{fig:FMNIST} shows the results of this experiment. One can observe that our \MnU~pruning criterion either with $\widetilde{\sigma}_{\widehat{w}_{j}}^{\mathrm{boot}}$ or $\widetilde{\sigma}_{\widehat{w}_{j}}^{\mathrm{pboot}}$ yielded a better (\ie, smaller) test accuracy drop compared to the $\ABS$ pruning criterion. The difference was rather small when the pruning level was relatively low, but the gap between the two increased as the pruning level increased. Using $\widetilde{\sigma}_{\widehat{w}_{j}}^{\mathrm{boot}}$ gave the most optimal result (defeated $\ABS$ 10 out of 11 times), but using $\widetilde{\sigma}_{\widehat{w}_{j}}^{\mathrm{pboot}}$ also gave better results than the $\ABS$ pruning criterion (defeated $\ABS$ 8 out of 11 times).

\subsection{Experiment 2: CIFAR-10}\label{section:experiment_2}
We further examined the performance of our \MnU~pruning criterion on the CIFAR-10 dataset~\citep{Krizhevsky2009}. Unlike Experiment 1, we utilized the whole training set, where 5,000 examples of the training set were used as validation set during training. We employed a smaller version of AlexNet~\citep{Krizhevsky2012}; the model has three convolution layers with ReLU activation function and max pooling, which yield feature maps of size 128*4*4 as output. The network also has four hidden layers with 512, 512, 512, and 256 neurons. As the rapid drop in test accuracy suggests, this model has relatively low degree of overparameterization. We chose cross entropy as loss function and the model was trained for 1,000 epochs with RMSprop. Due to large amount of computation required for $\widetilde{\sigma}_{\widehat{w}_{j}}^{\mathrm{boot}}$, we only used $\widetilde{\sigma}_{\widehat{w}_{j}}^{\mathrm{pboot}}$ with $B=200$. We empirically chose $\lambda^{*} = 1$.

We pruned parameters of each fully connected layer by using our \MnU~pruning criterion or the $\ABS$ pruning criterion; the pruning levels are shown in Figure~\ref{fig:CIFAR10}. After pruning, the model was retrained for 200 epochs. Figure~\ref{fig:CIFAR10} shows the results of this experiment. Our \MnU~pruning criterion with $\widetilde{\sigma}_{\widehat{w}_{j}}^{\mathrm{pboot}}$ gave a better test accuracy than the $\ABS$ criterion in 9 out of 11 times.

\subsection{Experiment 3: Overparameterization}\label{section:experiment_3}
To explore the case of high overparameterization, we used the MNIST dataset~\citep{LeCun1998MNIST} with a MLP model. The MLP model had three hidden layers with 512, 1024, and 512 neurons, respectively. The model was trained for 1,000 epochs with RMSprop. 

Due to the large amount of computation required for $\widetilde{\sigma}_{\widehat{w}_{j}}^{\mathrm{boot}}$, we only used $\widetilde{\sigma}_{\widehat{w}_{j}}^{\mathrm{pboot}}$ with $B=200$. We empirically chose $\lambda^{*} = 1$. For this experiment, we used the iterative pruning strategy described in~\citep{Han2015NIPS}. We pruned the parameters of each fully connected layer by using our \MnU~pruning criterion or the $\ABS$ pruning criterion with the pruning levels shown in Figure~\ref{fig:MNIST}. After pruning, the model was retrained for 200 epochs. Figure~\ref{fig:MNIST} shows the results of this experiment. Since the model is highly overparameterized, the performance drop was minimal, even when we pruned $99\%$ of the weights. In fact, pruning up to $95\%$ of the weights increased the test accuracy of the model, possibly because pruning decreased overfitting. Our \MnU~pruning criterion with $\widetilde{\sigma}_{\widehat{w}_{j}}^{\mathrm{pboot}}$ gave a better test accuracy than using the $\ABS$ pruning criterion in 10 out of 11 times.

Note that since the model was highly overparameterized, one could prune a large number of the weights without sacrificing too much predictive power. For example, Figure~\ref{fig:MNIST} shows that could prune $99\%$ of the weights while loosing only $0.33\%$ of the test accuracy (note that the unit of the $y$-axis in Figure~\ref{fig:MNIST} is basis point, and not percent). The performance difference between our \MnU~pruning criterion and the $\ABS$ pruning criterion might seem small due to this minimal test accuracy drop throughout all pruning levels. But when one takes this into account, one can see that our \MnU~pruning criterion still outperforms the $\ABS$ criterion consistently.

\begin{figure*}[h]
\begin{center}
\includegraphics[width=0.85\linewidth]{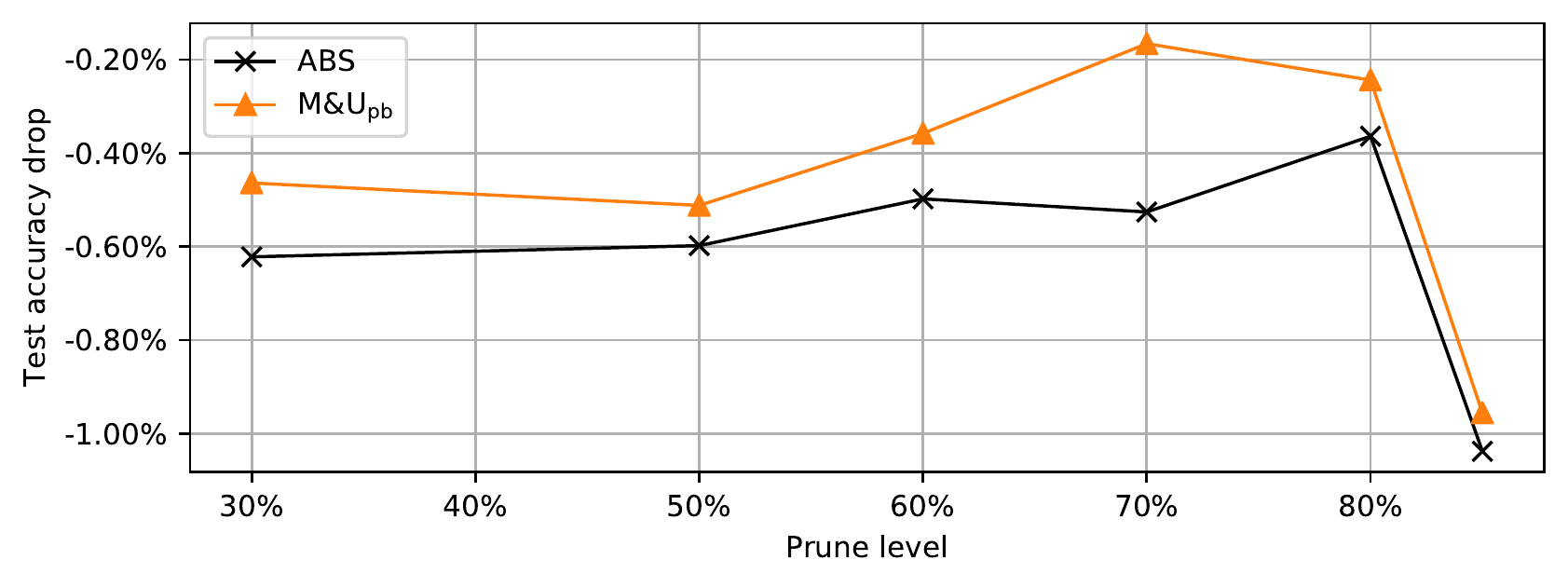}
\end{center}
\caption{\small Result of Experiment 4. Since the drop in test accuracy was at similar level across all the pruning levels we tested, a line plot is used. Orange triangles with `\MnU\textsubscript{pb}' mean that our \MnU~pruning method with $\widetilde{\sigma}_{\widehat{w}_{j}}^{\mathrm{pboot}}$ was used. Black crosses with `ABS' mean that the $\ABS$ pruning criterion from~\cite{Han2015NIPS} was used. The test accuracy of the unpruned model was $72.6820\%$.}
\label{fig:ImageNet}
\end{figure*}

\subsection{Experiment 4: ImageNet}\label{section:experiment_4}
ImageNet ILSCVR-12 dataset~\citep{ILSVRC15} has 1.28M training images, spread over 1.000 classes. To make 10 repetition of the experiment feasible, we used a subset that consists of 100 classes. We used VGG-11 ~\citep{Simonyan2014} architecture, shrank with the reduced number of classes. Like in Experiment 3, we used $\widetilde{\sigma}_{\widehat{w}_{j}}^{\mathrm{pboot}}$ with $B=200$. We empirically chose $\lambda^{*} = 10^{-3}$. We pruned the parameters of each fully connected layer by using our \MnU~pruning criterion or the $\ABS$ pruning criterion with the pruning levels shown in Figure~\ref{fig:ImageNet}. After pruning, the model was retrained for 40 epochs. Figure~\ref{fig:ImageNet} shows the results of this experiment. Our \MnU~pruning criterion with $\widetilde{\sigma}_{\widehat{w}_{j}}^{\mathrm{pboot}}$ gave a better test accuracy than using the $\ABS$ pruning criterion in 6 out of 6 times.

\section{Conclusion and Discussion}
Motivated by the Wald test statistic, we derived the \MnU~pruning criterion for neural networks. Our \MnU~pruning criterion takes both magnitude and uncertainty into account and can balance between them through the hyperparameter $\lambda$. It can be considered as a generalization of both the Wald test statistic and the magnitude based criterion~\citep{Han2015NIPS}. Our \MnU~pruning criterion is free from the scale variant issue, which the magnitude based pruning criterion suffer from. We also suggested a reparametrization of $\lambda$ that makes finding reasonable values for $\lambda$ easier. Further, we introduced `pseudo bootstrap', a very efficient method of measuring uncertainty of weights by tracking their updates during training. The experiments suggest that our \MnU~pruning criterion outperforms the criterion suggested by~\cite{Han2015NIPS} most of the time, in terms of test accuracy of the pruned model, with the best result achieved by using $\widetilde{\sigma}_{\widehat{w}_{j}}^{\mathrm{boot}}$. Thus, using our \MnU~pruning criterion can lead to models that require less memory and at smaller computational cost, while minimizing the loss of predictive power. In addition, Our MnU~pruning criterion can easily be applied to the pruning strategies that utilize the magnitude based pruning criterion.

In the future, we wish to investigate numerical and theoretical properties of pseudo bootstrap and compare it to that of the `usual' bootstrap. Due to limited amount of GPU resources, we had to keep our models relatively small. We would like to investigate the performance of our method on bigger and deeper neural network models. Further, it will be fruitful to combine our pruning criterion with other types of compression methods such as quantization, Huffman coding, and regularization.

\section*{Acknowledgment}
This work is supported by Google Cloud Platform through their research credits program.

We acknowledge support from the Danish Industry Foundation through the Industrial Data Analysis Service (IDAS).
\bibliographystyle{IEEEtran}
\bibliography{IEEEabrv,references/bibliography}

\end{document}